\documentclass[letterpaper, 10 pt, conference]{ieeeconf}  

\IEEEoverridecommandlockouts                              
\usepackage[utf8]{inputenc}
\usepackage[T1]{fontenc}

\usepackage{amsmath}
\usepackage{amsfonts}
\usepackage{amssymb}

\usepackage{amsthm}
\usepackage[ruled,vlined]{algorithm2e}
\usepackage{mathtools}
\usepackage{tabularx}
\usepackage{graphicx}
\usepackage{subfigure}
\usepackage{enumerate}
\usepackage{float}
\usepackage{url}
\usepackage{verbatim}
\usepackage{stfloats}
\usepackage{bbding}
\usepackage{multirow}

\usepackage[normalem]{ulem}
\usepackage{times}
\usepackage{xcolor}
\usepackage{cancel}

\newcommand{\delete}[1]{{\bgroup\markoverwith{\textcolor{red}{\rule[0.5ex]{2pt}{0.4pt}}}\ULon{#1}}}
\newcommand{\deletefig}[1]{{\bgroup\markoverwith{\textcolor{red}{\rule[2.5ex]{2pt}{2.0pt}}}\ULon{#1}}}

\makeatletter
\let\NAT@parse\undefined
\makeatother
\usepackage[colorlinks,linkcolor=red,anchorcolor=blue,urlcolor=magenta,citecolor=blue]{hyperref}

\usepackage{booktabs}
\usepackage{threeparttable}  
\usepackage{multirow}
\usepackage{pifont}

\usepackage{xcolor}
\definecolor{revised_color}{HTML}{00008B}

\usepackage{cite}

\graphicspath{{pics/}}

\title{\LARGE \bf
LF-VIO: A Visual-Inertial-Odometry Framework for Large Field-of-View Cameras with Negative Plane
}
\author{Ze Wang{$^{1,4}$}, Kailun Yang{$^{2}$}, Hao Shi$^{1}$,  Peng Li$^{1}$, {Fei Gao$^{3,4}$} and Kaiwei Wang$^{1}$
\thanks{*This was supported {in part by the National Natural Science Foundation of China (Grant No. 12174341),} in part by the Federal Ministry of Labor and Social Affairs (BMAS) through the AccessibleMaps project under Grant 01KM151112, in part by the University of Excellence through the ``KIT Future Fields'' project, in part by Hangzhou SurImage Technology Co. Ltd., and in part by Hangzhou HuanJun Technology Co. Ltd. {\textit{(Corresponding author: Kaiwei Wang.)}}}
\thanks{{$^{1}$State Key Laboratory of Modern Optical Instrumentation, Zhejiang University, China}}
\thanks{{$^{2}$Institute for Anthropomatics and Robotics, Karlsruhe Institute of Technology, Germany}}
\thanks{{$^{3}$State Key Laboratory of Industrial Control Technology, Zhejiang University, China}}
\thanks{{$^{4}$Huzhou Institute of Zhejiang University, Zhejiang University, China.}}
\thanks{{Email: \{wangze0527, haoshi, peng\_li, fgaoaa, wangkaiwei\}@zju.edu.cn, kailun.yang@kit.edu}}
}

\begin{document}

\maketitle
\thispagestyle{empty}
\pagestyle{empty}

\begin{abstract}
Visual-inertial-odometry has attracted extensive attention in the field of autonomous driving and robotics. The size of Field of View (FoV) plays an important role in Visual-Odometry (VO) and Visual-Inertial-Odometry (VIO), as a large FoV enables to perceive a wide range of surrounding scene elements and features. However, when the field of the camera reaches the negative half plane, one cannot simply use $\begin{bmatrix}u,v,1\end{bmatrix}^T$ to represent the image feature points anymore. To tackle this issue, we propose \emph{LF-VIO}, a real-time VIO framework for cameras with extremely large FoV. We leverage a three-dimensional vector with unit length to represent feature points, and design a series of algorithms to overcome this challenge. To address the scarcity of panoramic visual odometry datasets with ground-truth location and pose, we present the \emph{PALVIO} dataset, collected with a Panoramic Annular Lens (PAL) system with an entire FoV of $360^\circ{\times}(40^\circ{\sim}120^\circ)$ and an IMU sensor. With a comprehensive variety of experiments, the proposed LF-VIO is verified on both the established PALVIO benchmark and a public fisheye camera dataset with a FoV of $360^\circ{\times}(0^\circ{\sim}93.5^\circ)$. LF-VIO outperforms state-of-the-art visual-inertial-odometry methods. Our dataset and code are made publicly available at \url{https://github.com/flysoaryun/LF-VIO}
\end{abstract}

\section{Introduction}
\begin{figure}
\vskip-2ex
	\centering
	\includegraphics[width=1.0\linewidth]{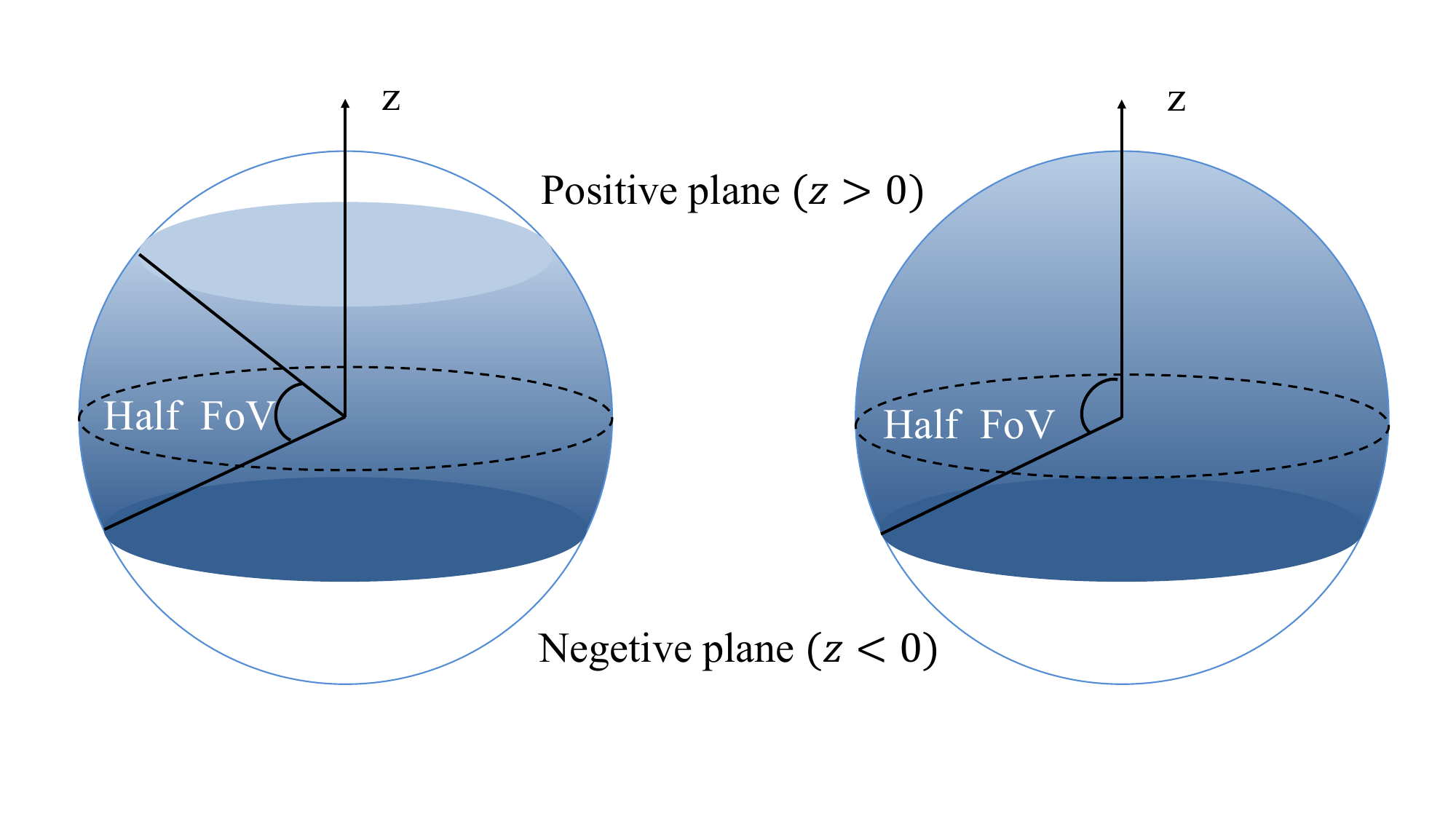}
	\vskip-7ex
	\caption{The left subfigure shows the half FoV of a panoramic annular camera or a catadioptric camera, and the right one shows the half FoV of a fisheye camera. The upper half above the dotted circle is the positive plane, while the bottom half is the negative plane.}
	\label{fig:negative_plane}
	\vskip-5ex
\end{figure}

With the rapid development of robotics and autonomous vehicles, Visual-Odometry (VO) and Visual-Inertial-Odometry (VIO) have been widely applied in navigation systems~\cite{chen2019palvo,hu2019indoor,seok2019rovo,seok2020rovins}.
{Meanwhile, cameras with a large Field of View (FoV) have been increasingly put into integration in such systems to enable a wide range of surrounding sensing of scene elements and features, which are often beneficial for upper-level vision perception and odometry tasks~\cite{yang2020dspass,lin2018pvo,gao2021lovins,chen2021semantic,yang2021capturing,jaus2021panoramic,yang2020omnisupervised,shi2022panoflow}.}
Some modern panoramic cameras~\cite{sun2019multimodal,yang2019pass,shan2021lvi} even have a negative plane field, that facilitates an ultra-wide surrounding understanding, where imaging points may appear on the negative plane ($z\textless 0$, see Fig.~\ref{fig:negative_plane}).

Nowadays, there are many VIO frameworks~\cite{forster2014svo,qin2018vins,campos2021orb} that support different kinds of camera models including pinhole-, fisheye-, and the omnidirectional camera model introduced by Scaramuzza~\textit{et al.}~\cite{scaramuzza2006toolbox}.
While some camera models, \textit{e.g.,} the model introduced by Scaramuzza~\textit{et al.}, can support to work with cameras with a negative semi-planar imaging region, all of existing systems discard the points within the negative half plane during the subsequent processing,
due to the usage of $\begin{bmatrix}u,v,1\end{bmatrix}^T$ to represent the location of image feature points.
When the field of the camera reaches the negative half plane, one cannot simply use this representation anymore.
This is a severe issue as such a negative semi-planar region may exist in various large-FoV systems like fisheye-, panoramic annular-, and catadioptric cameras, while discarding the features leaves abundant yet important features unused.
Besides, $u$ and $v$ will increase rapidly near $180^\circ$, which is disadvantageous to some following algorithms such as PnP and epipolar constraints for solving the rotation matrix $R$ and the translation vector $T$, leading to degraded tracking accuracy and even failures.

To tackle this challenge, we propose \emph{LF-VIO}, a real-time VIO framework for cameras with very large FoV.
In LF-VIO, we adapt a KLT sparse flow method~\cite{lucas1981iterative} to extract feature points. We then propose to use a feature point vector with unit length to represent the features. And the RANSAC method is then used to pick out outliers.
Specifically, in the process of initialization, the epipolar geometry is used to initialize if two frames have large enough parallax.
After decomposing the essential matrix into rotation matrix and translation vector, the correct rotation matrix and translation vector are selected.
Then, triangulation and EPnP alternation methods~\cite{lepetit2009epnp} are introduced to initialize the depth of feature points and every pose in the sliding window,
and a tightly coupled optimization method is used to solve all the rotation and translation matrices given by the sliding window.
After vision initialization, the IMU data and image data are aligned to recover the scale information.
In our approach, we take the visual re-projection error, the IMU pre-integration error, and the marginalization error to solve the optimization problem.

\begin{figure}[t]
	\centering
	\includegraphics[width=1.0\linewidth]{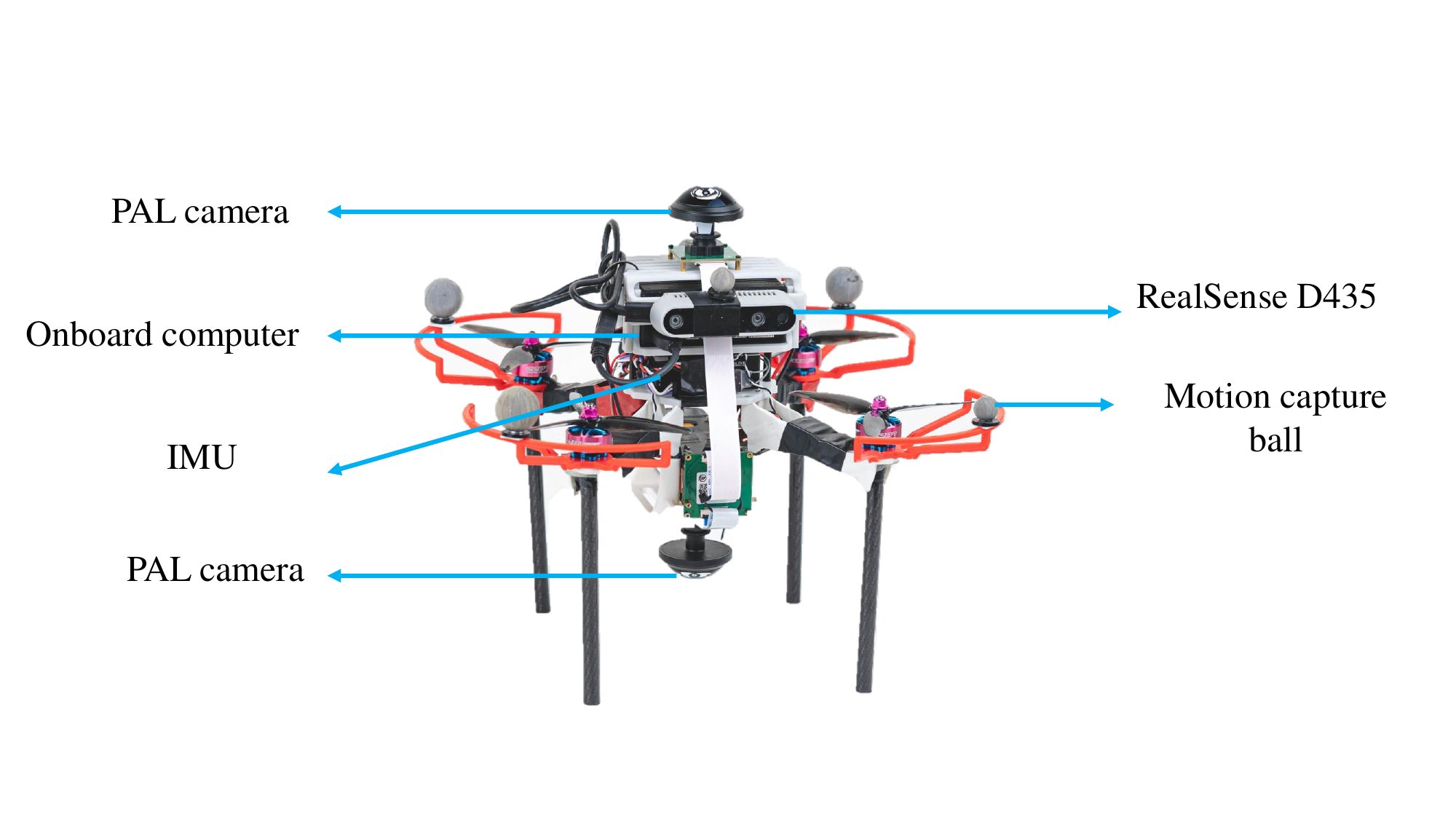}
	\vskip-6ex
	\caption{Our experiment platform has two Panoramic Annular Lens (PAL) cameras, a RealSense D435 sensor, a flight control system with an IMU sensor and an onborad computer.}
    \label{fig:platform}
    \vskip-4ex
\end{figure}

To address the scarcity of panoramic visual odometry datasets with ground-truth location and pose,
we introduce the \emph{PALVIO} dataset,
which is collected via a Panoramic Annular Lens (PAL) system with an entire FoV of $360^\circ{\times}(40^\circ{\sim}120^\circ)$, an IMU sensor, and a motion capture device (see Fig.~\ref{fig:platform}).
We conduct extensive quantitative experiments to verify our proposed LF-VIO framework on both the established PALVIO benchmark and a public fisheye camera dataset~\cite{shan2021lvi} with a FoV of $360^\circ{\times}(0^\circ{\sim}93.5^\circ)$.
Our comprehensive variety of investigations on different FoVs confirms that the information of the negative plane are very important for a VIO system.
The proposed method clearly outperforms state-of-the-art VIO frameworks.
In addition, we show that our method is also beneficial when it is integrated with a LiDAR-Visual-Inertial Odometry~\cite{shan2021lvi}.
Furthermore, since feature extraction is performed on the raw panoramic image, bypassing the complex panorama unfolding and the step of cropping the unfolded panorama into $4{\sim}5$ pinhole sub-images, LF-VIO has a fast speed, which is suitable for real-time mobile robotics applications.

In summary, we deliver the following contributions:
\begin{itemize}
    \item We propose \emph{LF-VIO}, a visual-inertial-odometry framework for cameras with a large field of view.
    \item We present a robust initialization method and algorithmic adjustments to support the feature points on the negative half plane.
    \item We create and release the \emph{PALVIO} dataset, which is collected with a Panoramic Annular Lens (PAL) system and an IMU sensor with the ground-truth location and pose obtained via a motion capture device.
    \item LF-VIO outperforms state-of-the-art visual-inertial-odometry methods on multiple wide-FoV datasets and experiment sequences with a negative plane.
\end{itemize}

\section{Related Work}
In this section, a brief review of representative works is presented on visual inertial odometry and panoramic SLAM (Simultaneous Localization And Mapping) frameworks.
\subsection{Visual inertial odometry}
Visual-Inertial-Odometry (VIO) is the process of a state estimator using vision and inertial measurement unit estimating the 3D pose (local position and orientation) and velocity relative to a local starting position.
Qin~\textit{et al.}~\cite{qin2018vins} proposed VINS, a robust and universal monocular vision inertial state estimator.
Recently, the Semi-direct Visual-inertial Odometry (SVO2.0)~\cite{forster2014svo} has been released for monocular and multi-camera systems.
Campos~\textit{et al.}~\cite{campos2021orb} proposed ORB-SLAM3, an accurate open-source library for vision, visual inertial, and multi-map SLAM.
LiDAR-visual-inertial sensor fusion frameworks have also been developed in recent years such as LVI-SAM~\cite{shan2021lvi} and R3LIVE~\cite{lin2021r3live}.
In addition to the feature-point-based methods, there are also direct methods for VO and VIO, which tend to be more sensitive to light than the feature point methods.
Engel~\textit{et al.}~\cite{engel2017dso} proposed Direct Sparse Odometry (DSO) using a fully direct probabilistic model and Wang~\textit{et al.}~\cite{wang2017stereo_dso} combined it with inertial systems and stereo cameras.
However, most of the works do not directly work well with panoramic camera systems  which capture $360^\circ$ omnidirectional contents.
Different from these previous works, we address panoramic visual inertial odometry for cameras with extremely large field of view and a negative plane.
 
\subsection{Panoramic SLAM frameworks}
Panoramic SLAM systems often leverage fisheye cameras, panoramic annular cameras, and multi-camera systems.
Sumikura~\textit{et al.}~\cite{sumikura2019openvslam} proposed OpenVSLAM, which is a VO framework supporting panoramic cameras.
Wang~\textit{et al.}~\cite{wang2018cubemapslam} proposed CubemapSLAM, a piecewise-pinhole monocular fisheye SLAM system.
Seok and Lim~\cite{seok2020rovins} introduced ROVINS, which uses four fisheye cameras.
Chen~\textit{et al.}~\cite{chen2019palvo,chen2021pa_slam} tackled VO with a panoramic annular camera.
More recently, Wang~\textit{et al.}~\cite{wang2022pal_slam} proposed PAL-SLAM, which supports panoramic annular cameras.
However, the FoV of their used panoramic camera is only $360^{\circ}{\times}(45^{\circ}{\sim}85^{\circ})$, which does not have a negative half plane.
Moreover, based on different sensor types, many other odometry estimation methods emerge in the field, including fisheye-based~\cite{heng2016semi_dso_fisheye,liu2017direct_fisheye,matsuki2018omnidirectional_dso,wang2018realtime_omnidirectional_semi_dense,ramezani2018pose_estimation_omnidirectional}, multi-camera-based~\cite{liu2018towards_multicamera,jaramillo2019visual_single_stereo,yogamani2019woodscape,seok2019rovo,won2020omnislam,kumar2021omnidet}, and systems that combine omnidirectional imaging with other sensing units like GPS~\cite{yu2019gps} and LiDAR sensors~\cite{kang2021rpv_slam,xiang2019vilivo}.

Some of the above algorithms have already completed the panoramic image unfolding step, and some algorithms use the direct method or the semi-direct method, which are greatly affected by the changes of illumination.
While SVO2.0~\cite{forster2014svo} and VINS~\cite{qin2018vins} support panoramic annular camera and fisheye camera, both of them convert the model into a pinhole camera model, which makes it impossible to correctly use the important negative half-plane information. 
Thereby, all the aforementioned VIO works do not support a panoramic annular camera or a catadioptric camera very well.
Furthermore, most of the systems that support panoramic annular cameras are still VO frameworks~\cite{chen2019palvo,yang2020dspass,chen2021pa_slam,wang2022pal_slam}.
To fill this gap, we propose a VIO framework to support all types of large-FoV cameras with a negative half plane.

\section{LF-VIO: Proposed Framework}

In this section, we describe in detail the proposed LF-VIO framework for large-FoV panoramic cameras with a negative plane. LF-VIO mainly includes three parts: tracking, initialization, and back-end optimization, which will be introduced in the following subsections.
          
\subsection{Tracking}
In order to obtain the relationship of different image frames, we use a tracking method to establish correspondences between two frames.
First, we extract Shi-Tomasi corners~\cite{shi1994good} and use the Lucas-Kanade method~\cite{lucas1981iterative} to track these corners.
If the corners are out of the border or the quality is lower than a threshold, we will eliminate these points.
We then convert these points to a feature vector using a corresponding camera model.

The camera model represents the mathematical relationship between the pixel coordinates on the camera and the image plane.
Due to the large FoV, the imaging projection process of a panoramic camera, \textit{e.g.,} the panoramic annular optical system~\cite{chen2019palvo}, is significantly different from the pinhole camera model.
It is necessary to introduce a model suitable for large-FoV imaging which modifies from existing models.
For the imaging system, the mapping relationship between the camera coordinate system and the pixel coordinate system can be obtained by calibrating the intrinsic parameters, and the projection relationship between them can be described by the unified spherical model, which is a unit sphere with the camera as its origin. Each point of the pixel coordinate will correspond to a vector whose spherical center points to the unit sphere.
For an omnidirectional system, we set the camera center $C$ on the Z-axis and take the quadric center $C$ as the origin.
For a space point $P(x,y,z)$ that is projected onto the unit sphere, the projection point $P_s$ is obtained:
\begin{equation}
P_s=\frac{v}{||v||_2},
\end{equation}
\begin{equation}
v=\begin{bmatrix} x_m\\y_m\\a_0+a_1\rho+a_2\rho^2+...a_n\rho^ N\end{bmatrix}, \\
\rho=\sqrt{x_m^2+y_m^2},    
\end{equation}
where the coefficients $a_i\left(i=1,2...N\right)$ are the polynomial parameters and $x_m,y_m$ denotes the location of a pixel point.
These coefficients are given via a panoramic camera calibration process with the OmniCalib calibration toolbox~\cite{scaramuzza2006toolbox}.

Since some panoramic annular cameras and fisheye cameras can see the object behind the camera, that is, the negative half plane of the image $(z<0)$,
we propose to use $\begin{bmatrix}\alpha,\beta,\gamma\end{bmatrix}^T$ to represent the scene spot vector:
\begin{equation}
\begin{bmatrix} \alpha \\\beta \\\gamma\end{bmatrix}={P_s=}\pi^{-1}_s\left( \begin{bmatrix} x_m\\y_m\end{bmatrix}\right),
\end{equation}
\begin{equation}
\alpha^2+\beta^2+\gamma^2=1,
\end{equation}
where $\pi_s$ means the mapping relationship from unit 3D coordinate point to pixel point.
We can also use the MEI~\cite{mei2007single} or the Kannala-Brandt~\cite{kannala2006generic} model, which can also support negative plane to represent this relationship.
{On the PALVIO dataset, we choose the camera model from Scaramuzza~\textit{et al.}~\cite{scaramuzza2006toolbox} for efficiency and distortion considerations.}
Afterwards, a series of algorithms need to be adjusted to adapt to work with a camera which has a negative plane.

\begin{figure}[t]
	\centering
	\includegraphics[width=1.0\linewidth]{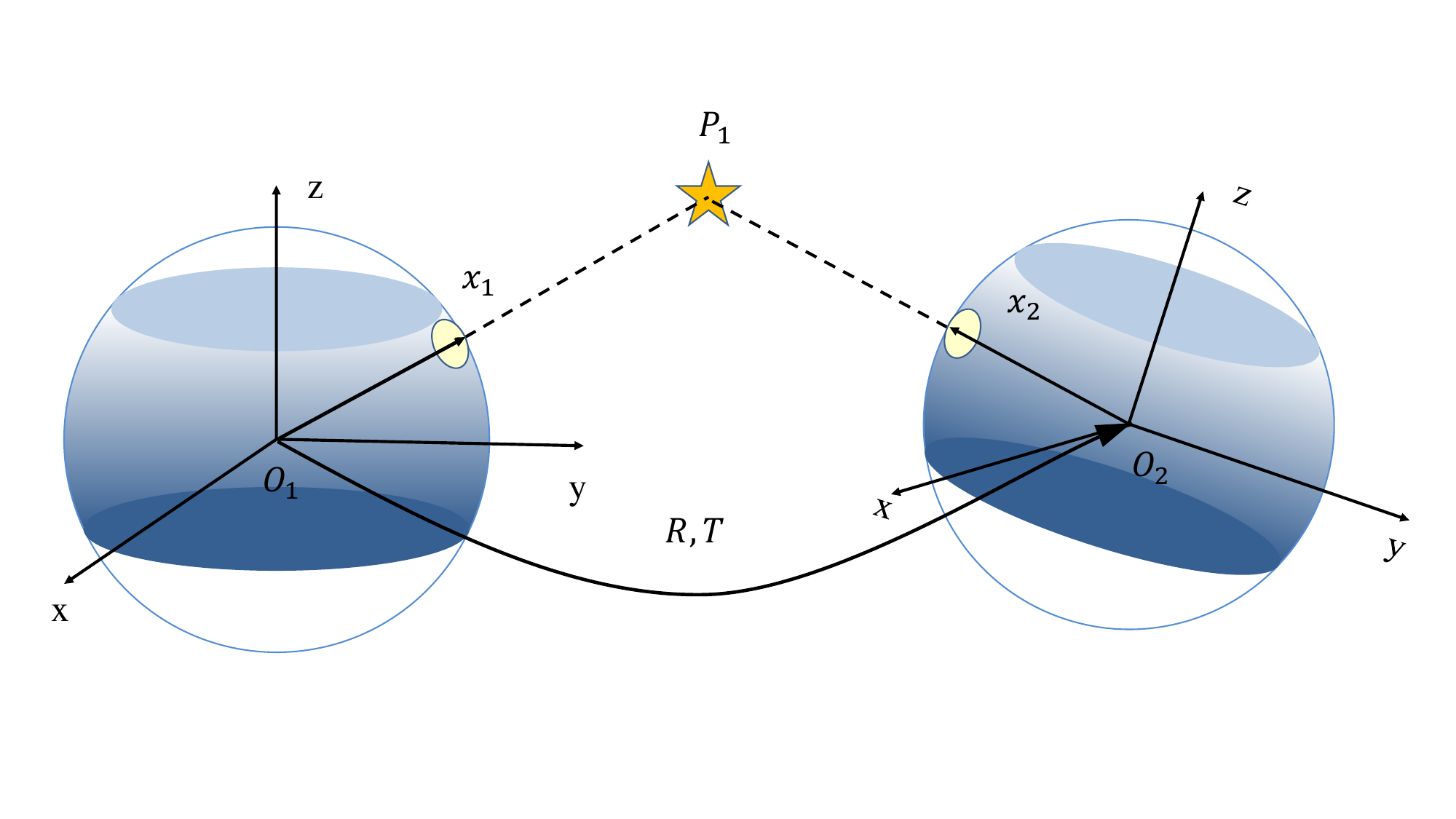}
	\vskip-8ex
	\caption{The epipolar constraint. The vectors of two corresponding features and their rotation and translation are on the same plane.}
	\label{fig:epipolar}
\vskip-2ex
\end{figure}

In the original image, an optical flow method (KLT sparse flow)~\cite{lucas1981iterative} is leveraged to track the feature points, and the polar geometry RANSAC method is used to eliminate the outlier points:
\begin{equation}
{
 x_2^T\left( \left[T\right]^\wedge R \right)x_1=0
},
\end{equation}
\begin{equation}
x_1=\begin{bmatrix} \alpha_1 \\\beta_1 \\\gamma_1\end{bmatrix},x_2=\begin{bmatrix} \alpha_2 \\\beta_2 \\\gamma_2\end{bmatrix},
\end{equation}
where $x_1$ and $x_2$ are the unit vectors corresponding to their space points $P_1$ and $P_2$ in Fig.~\ref{fig:epipolar}, {and the notation $[T]^{\wedge}$ denotes the skew-symmetric cross
product matrix of $T\in{\mathbb{R}^3}$.}

\subsection{Initialization}
\begin{figure}[t]
	\centering
	\includegraphics[width=1.0\linewidth]{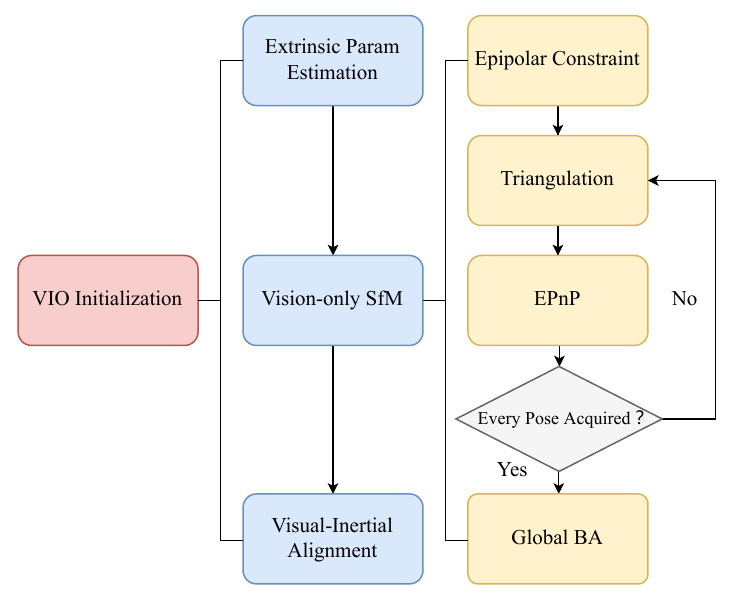}
	\vskip-2ex
	\caption{The overview of the LF-VIO initialization.}
	\label{fig:LF-VIO initialization}
\vskip-4ex
\end{figure}

The initialization procedure, as shown in Fig.~\ref{fig:LF-VIO initialization}, includes the three following parts: $R_c^b$ estimation, Vision-only Structure from Motion (SfM), and Visual-inertial alignment.

\noindent$\mathbf{R_c^b}$ \textbf{estimation:}
$R_c^b$ and $q_c^b$ means the rotation matrix and quaternion from the camera coordinate to the IMU coordinate.
We first maintain several frames in a sliding window to keep a low computational complexity.
From frame $k$ and frame $k+1$, we can obtain the following equation:
\begin{equation}
q_b^c\otimes q_{b_{k+1}}^{b_{k}}=q_{c_{k+1}}^{c_{k}}\otimes{q_b^c},
\end{equation}
where $q^{b_k}_{b_{k+1}}$ is computed with IMU pre-integration and $q^{c_k}_{c_{k+1}}$ is computed with epipolar constraint and essential matrix decomposition. From this equation, we can estimate $q_b^c$, and obtain $R_c^b$ according to the commonly used quaternion to rotation matrix algorithm.

\begin{figure}[!t]
	\centering
	\includegraphics[width=1.0\linewidth]{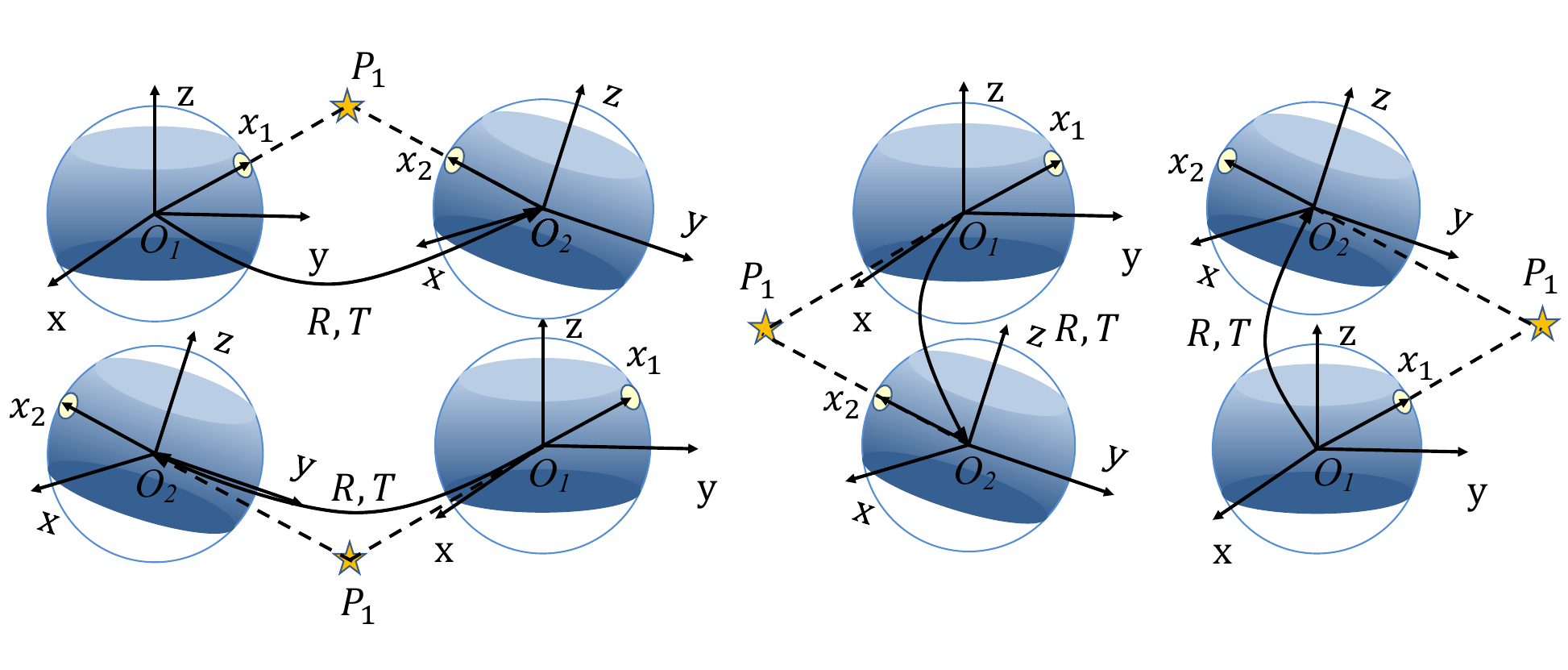}
	\vskip-2ex
	\caption{Essential matrix decomposition. There are four sets of solutions to the essential matrix decomposition, but only the first set of solutions is in line with the actual situation.}
	\label{fig:E}
\vskip-4ex
\end{figure}

\noindent\textbf{Vision-only SfM:}
We first search for two frames which have a large parallax.
Then, we use the epipolar constraint to acquire the essential matrix.
Subsequently, four sets of solutions are obtained by decomposing the essential matrix.
At this time, we use the method depicted in the following equation, to choose the right $R$ and $T$, so that more feature point pairs satisfy the requirement that the dot with the landmarks are always positive in Fig.~\ref{fig:E}:
\begin{equation}
    \overrightarrow{O_1P_1}\cdot \overrightarrow{x_1}>0\ \&\&\ \overrightarrow{O_2P_1}\cdot \overrightarrow{x_2}>0.
\end{equation}
We then use the obtained $R$ and $T$ to triangulate landmarks $\begin{bmatrix} x_l,y_l,z_l\end{bmatrix}^T$:
\begin{equation}
\lambda\begin{bmatrix} \alpha\\\beta\\\gamma\end{bmatrix}=R\begin{bmatrix} x_l\\y_l\\z_l \end{bmatrix}+T.
\end{equation}
According to the landmarks, the EPnP method~\cite{lepetit2009epnp} is introduced to obtain the $R$ and $T$ between the initialization frames.
After that, triangulation and EPnP methods are used alternately.
By analogy, the three-dimensional spatial positions of every $R$ and $T$ and landmarks in the sliding window are acquired.

Finally, the re-projection error method is used to optimize the $R$ and $T$ of all frames in the sliding window:
\begin{equation}
\underset{R,T}{min}\left\{ \sum_{i=1}^M\sum_{j=1}^N||\begin{bmatrix} \alpha_i\\\beta_i\\\gamma_i\end{bmatrix}-\frac{ RP_j+T}{||R P_j+T||_2}||^2\right\},
\end{equation}
where $M$ is the number of observations and $N$ is the number of landmarks.

\begin{figure}[h]
	\centering
	\includegraphics[width=1.0\linewidth]{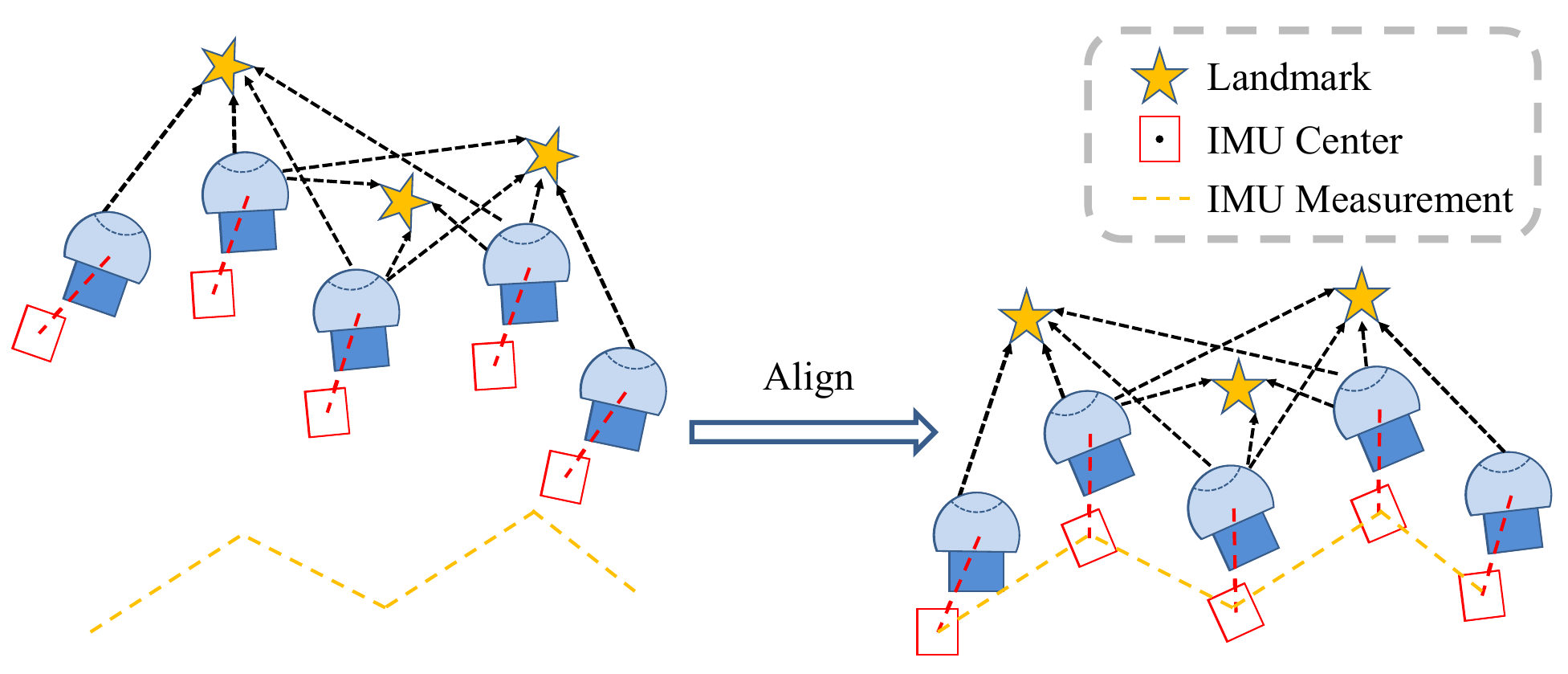}
	\vskip-2ex
	\caption{IMU and visual alignment. The up-to-scale visual structure is aligned with the pre-integrated IMU measurements.
	}
	\label{fig:align}
	\vskip-4ex
\end{figure}

\noindent\textbf{Visual-inertial alignment.}
An illustration of the visual-inertial alignment is shown in Fig.~\ref{fig:align}.
We match the up-to-scale visual structure with IMU pre-integration.
Then, we calibrate the gyroscope bias, initialize velocity, gravity vector and metric scale, and finally align the gravity direction with the $Z$ axis.
{Because the feature vector is expressed by a unit vector, we innovatively design an initialization method and adjust a series of algorithms regarding epipolar constraint, triangulation, and re-projection error.}

\subsection{Tightly coupled LF-VIO}
After estimation, we use a sliding window-based tightly-coupled monocular odometry for state estimation.
The state vector is defined as: 
\begin{equation}
\begin{aligned}
    \chi & =[x_,…,x_{N},{T_{c^0}^B,…,T_{c^N}^B},\lambda_{d_1},…\lambda_{d_m}],\\
     x_k & =[P_{B_K}^W,V_{B_K}^W,q_{B_K}^W,b_a,b_g],\\
     t_{c^i}^B &= [{P_{c^i}^B},q_{c^i}^B],
\end{aligned}
\end{equation}
where $\lambda_{d_m}$ is the inverse distance of the $m_{th}$ feature from its first observation to the unit sphere.
$N$ is the total number of sliding windows.
The $b_a$ and $b_g$ are the basis of accelerometer and gyroscope.
During the optimization, the calibration parameter {$T_{c^N}^B$} will be updated and converge to a reasonable value.
The optimization only considers the IMU measurement, the visual measurement, and the marginalization residual:
\begin{equation}
\begin{aligned}
\min\limits_{\chi}
\Bigg\{
||\mathbf{r}_p-\mathbf{H}_p\chi||^2
+&\sum_{k\in\mathcal{B}}||{r_\mathcal{B}\left(\hat{z}_{b_{k+1}}^{b_k},\chi\right)||^2_{\mathbf{P}^{b_k}_{b_{k+1}}}}\\
+&\sum_{\left(i,j\right)\in\mathbf{C}}||r_\mathbf{C}\left(\hat{z}_{l}^{c_j},\chi\right)||^2_{\mathbf{P}^{c_j}_{l}}
\Bigg\},
\end{aligned}
\end{equation}
where $r_\mathcal{B}\left(\hat{z}_{b_{k+1}}^{b_k},\chi\right)$ and $r_\mathbf{C}\left(\hat{z}_{l}^{c_j},\chi\right)$ are IMU measurement residual and camera measurement residual, respectively.
$\textbf{r}_p,\textbf{H}_p$ are the prior information.

\noindent\textbf{Visual measurement residual:}
Considering that our addressed large-FoV panoramic cameras have a negative plane, we use the unit sphere to define our visual residual.
At this time, $\lambda_d$ represents the inverse of the distance from the unit to the feature point {and the $l$th feature is observed in the $i$th image}, and we define the visual measurement residual as:
\begin{equation}
\textbf{r}_c\left( \hat{z}_l^{c_j},\chi \right)=\left[
    \begin{array}{cc}
\textbf{b}_1 & \textbf{b}_2
    \end{array}
\right]^T \cdot\left( \hat{\bar{\mathcal{P}}}_l^{c_j}-\frac{\mathcal{P}_l^{c_j}}{||\mathcal{P}_l^{c_j}||}\right),
\end{equation}
\begin{equation}
\begin{aligned}
{\mathcal{P}}_l^{c_j} = & \textbf{R}_b^c\bigg( \textbf{R}_w^{b_j}\bigg( \textbf{R}_{b_i}^w \bigg(  \textbf{R}_{c}^{b} \frac{1}{\lambda_d}\pi^{-1}_s\bigg({\begin{bmatrix}
\hat{\bar{x}}_l^{c_i}\\\hat{\bar{y}}_l^{c_i}
\end{bmatrix}}\bigg)\\
& + {\textbf{P}_{c}^b} \bigg)+{\textbf{P}_{b_i}^w}-{\textbf{P}_{b_j}^w} \bigg)-{\textbf{P}_c^b} \bigg),
\end{aligned}
\end{equation}
where $\textbf{b}_1$ and $\textbf{b}_2$ are two arbitrarily selected orthogonal
bases which span the tangent plane of $\hat{\bar{\mathcal{P}}}_l^{c_j}$.
Ceres Slover~\cite{ceres} is used to solve the nonlinear maximum posterior estimation problem. We use the Huber loss function to reduce the influence outliers for better system robustness.

\section{Experiments}
\subsection{Datasets}
\noindent\textbf{PALVIO dataset.}
To address the lack of panoramic visual odometry dataset with ground-truth location and pose, we collect and release our PALVIO dataset.
PALVIO is collected with two panoramic annular cameras (see Fig.~\ref{fig:platform}), an IMU sensor, and a RealSense D435 sensor with ground truth under motion capture in a computing platform with a quad-core Intel i7-8550U processor.
The panoramic cameras capture monocular $1280{\times}960$ images at $30Hz$, with a FoV of $360^\circ{\times}(40^\circ{\sim}120^\circ)$.
The IMU provides the angular velocity and acceleration at $200Hz$ and the motion capture system provides the position and attitude at $10Hz$.
We collect $10$ sequences in an area of of $8m{\times}10m$ indoors (ID).
For the vision data, the data captured by the stereo panoramic cameras and the RealSense camera are both made publicly available. In this work, we mainly use the data captured by the top panoramic camera for experiments.

\noindent\textbf{LVI-SAM dataset.} We use a public fisheye dataset LVI-SAM~\cite{shan2021lvi} to verify the generalizability of our approach.
The data collection sensor suite includes: a Velodyne VLP-16 LiDAR sensor, an FLIR BFS-U3-04S2M-CS camera, a MicroStrain 3DM-GX5-25 IMU sensor, and Reach RS+ GPS. $\emph{Jackal}$- and $\emph{Handheld}$ datasets are gathered by an unmanned ground vehicle.
The FoV of their used fisheye camera is approximately $360^\circ{\times}(0^\circ{\sim}93.5^\circ)$.
We use the LVI-SAM dataset together with their GPS measurements, which are treated as the ground truth for evaluation.

\begin{table*}[!t]
 \setlength{\tabcolsep}{2.2pt}
	\centering
	\begin{threeparttable}
	\caption{Accuracy analysis of LF-VIO using images with different field of view on the PALVIO ID06 set.}
	\label{tab:fov}
\renewcommand\arraystretch{1.4}{\setlength{\tabcolsep}{7.5mm}{\begin{tabular}{cccccc}
\toprule
Field of View & $40^{\circ}{\sim}120^{\circ}$ & $40^{\circ}{\sim}110^{\circ}$ & $40^{\circ}{\sim}100^{\circ}$ & $40^{\circ}{\sim}90^{\circ}$ & $40^{\circ}{\sim}80^{\circ}$ \\
\midrule
\midrule
RPEt (\%) &\textbf{2.814} & 3.097 & \underline{2.923} & 2.933 & 3.392 \\
RPEr (degree/m) & \underline{0.397} & \textbf{0.393} & 0.402 & 0.433 & 0.631 \\
ATE (m) &\textbf{0.093}& 0.182 & \underline{0.112} & 0.124 & 0.171 \\

\midrule
Field of View & $50^{\circ}{\sim}120^{\circ}$ & $60^{\circ}{\sim}120^{\circ}$ & $70^{\circ}{\sim}120^{\circ}$ & $80^{\circ}{\sim}120^{\circ}$ & $90^{\circ}{\sim}120^{\circ}$ \\
\midrule
RPEt (\%) 
&\textbf{2.585}&\underline{2.712}&3.584&3.525&5.223
\\
RPEr (degree/m) 
&\textbf{0.394}&\underline{0.398}&0.417&0.449&0.490
\\
ATE (m) 
&\textbf{0.081}&\underline{0.117}&0.250&0.252&0.475
\\
	\bottomrule
	\end{tabular}}}
	\end{threeparttable}
\vskip-4ex
\end{table*}

\subsection{Investigation on Different Field-of-View}
\begin{figure}[t]
	\centering
	\includegraphics[width=0.7\linewidth]{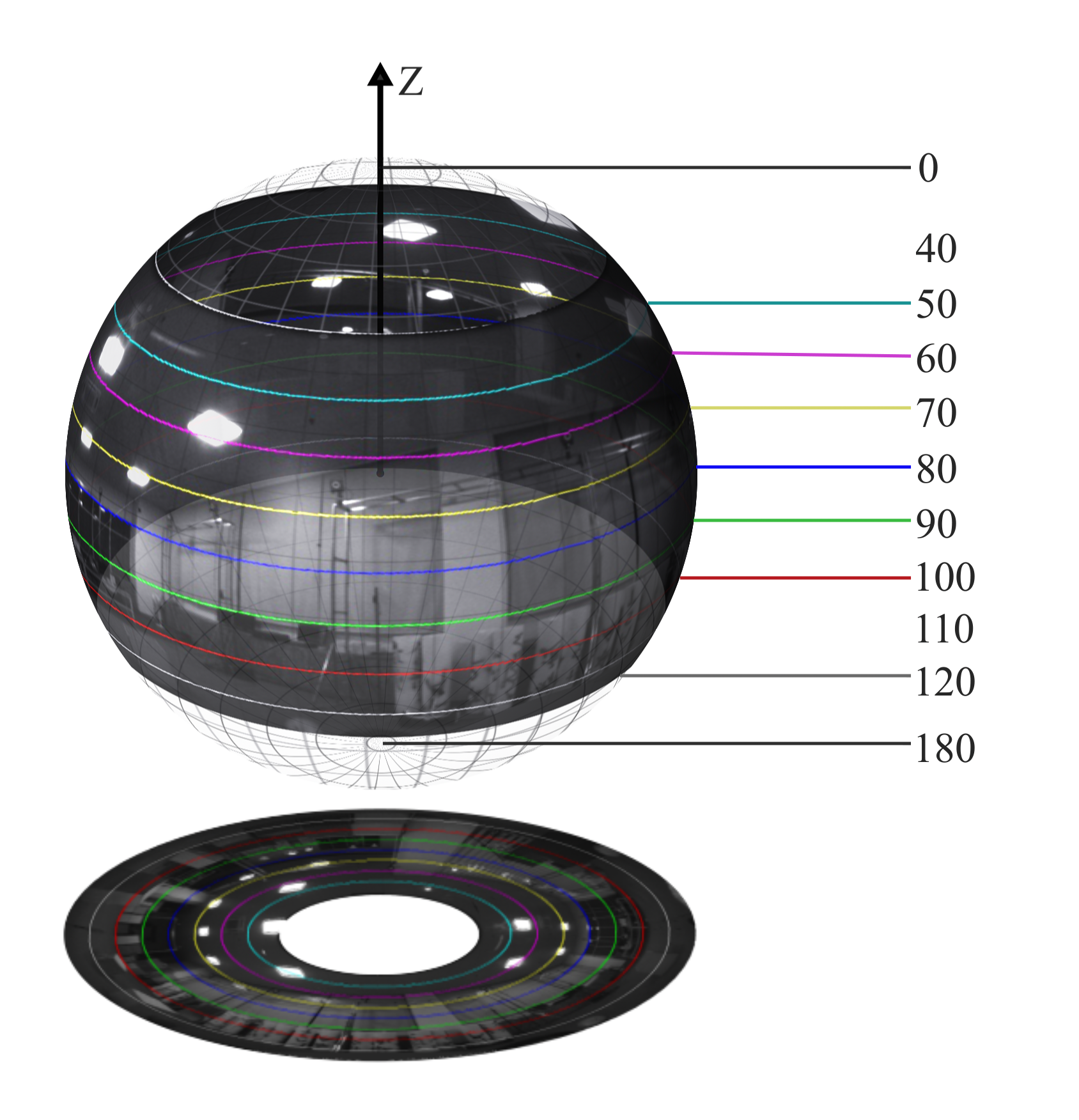}
	\vskip-3ex
	\caption{{Variation of FoVs with a step of $10^\circ$ using the annular images.}}
	\label{fig:different_fov}
	\vskip-4ex
\end{figure}

To validate the importance of the information of the negative plane, we study the influence of using inputs with different FoVs, by only extracting features from the corresponding angle range. 
First, we gradually decrease the FoV from $40^{\circ}{\sim}120^{\circ}$ to $40^{\circ}{\sim}80^{\circ}$, with a variation step of $10^\circ$ (see Fig.~\ref{fig:different_fov}), and perform the experiment using our LF-VIO framework on the PALVIO ID06 set, which is chosen as a representative sequence, where the results are shown in Table~\ref{tab:fov}.
It can be seen that using the entire FoV of the panoramic camera ($40^{\circ}{\sim}120^{\circ}$) leads to the best performances in RPEt and ATE.
When the FoV reduces to only the positive plane, \textit{i.e.}, in the situations of $40^{\circ}{\sim}90^{\circ}$ and $40^{\circ}{\sim}80^{\circ}$, the performance degrades dramatically, verifying the importance of the information from the negative plane.

\begin{figure*}[h]
	\centering
	\subfigure[Top Trajectory]{
		\includegraphics[width=0.48\textwidth]{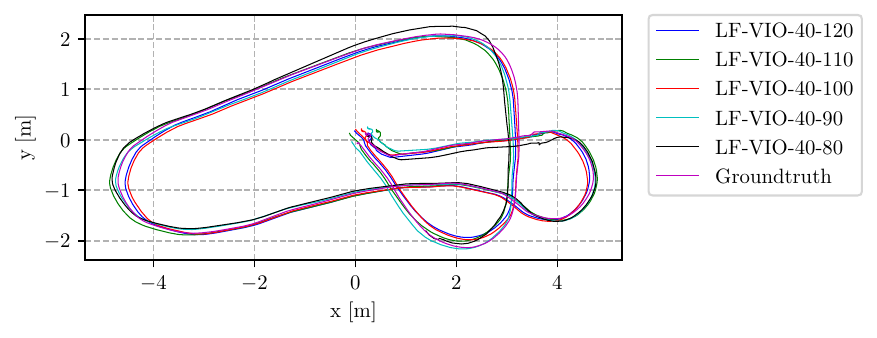}
		\label{fig:up_traj}
	}
	\subfigure[Top Trajectory]{
		\includegraphics[width=0.48\textwidth]{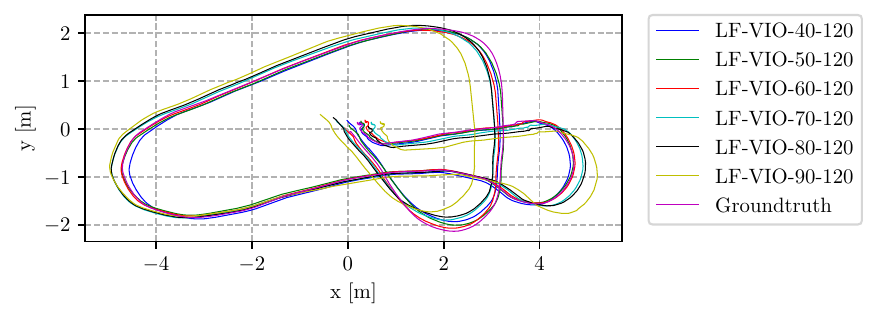}
		\label{fig:down_traj}
	}
	\subfigure[Translation and Rotation Error]{
		\includegraphics[width=0.48\textwidth]{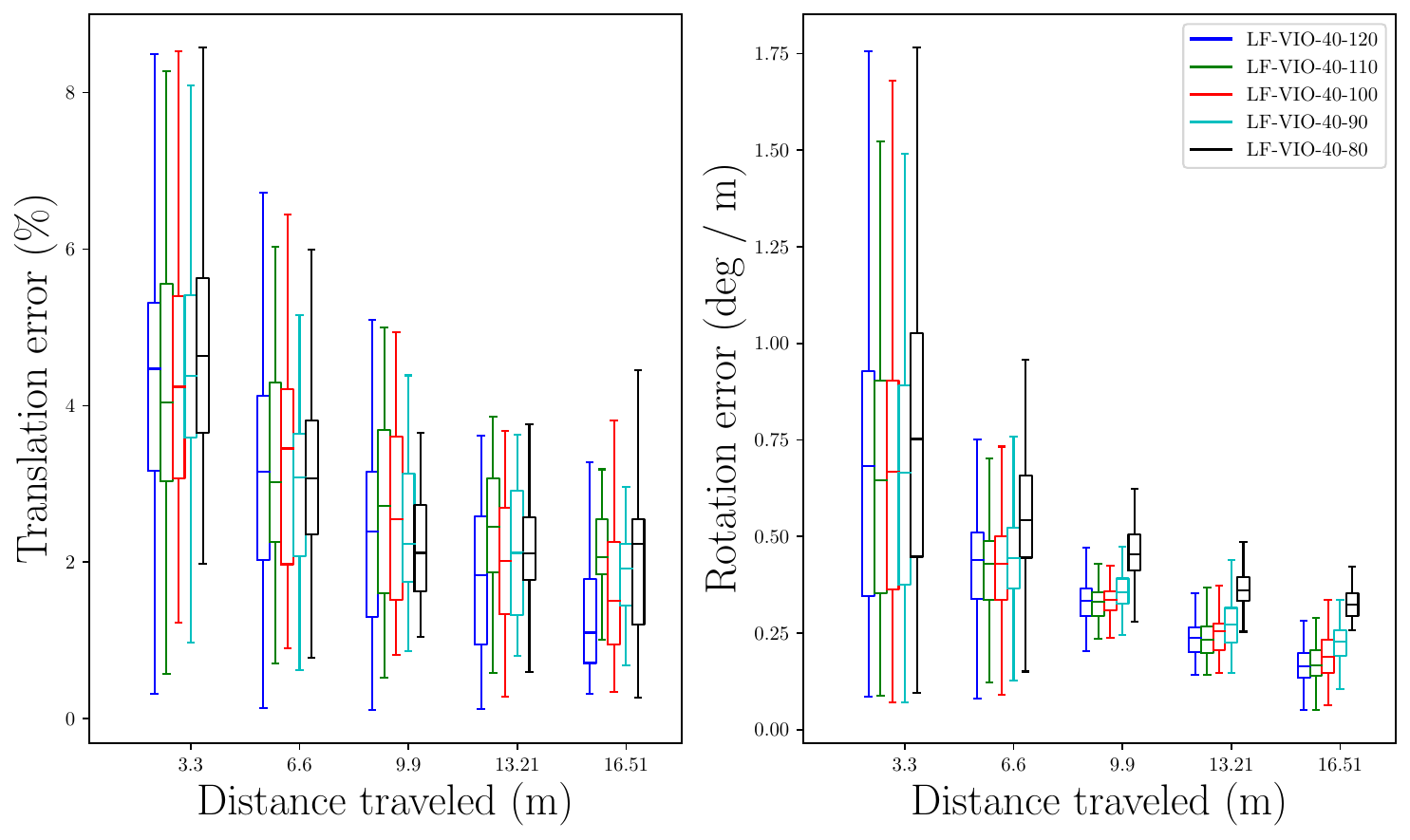}
		\label{fig:up_error}
	}
	\subfigure[Translation and Rotation Error]{
		\includegraphics[width=0.48\textwidth]{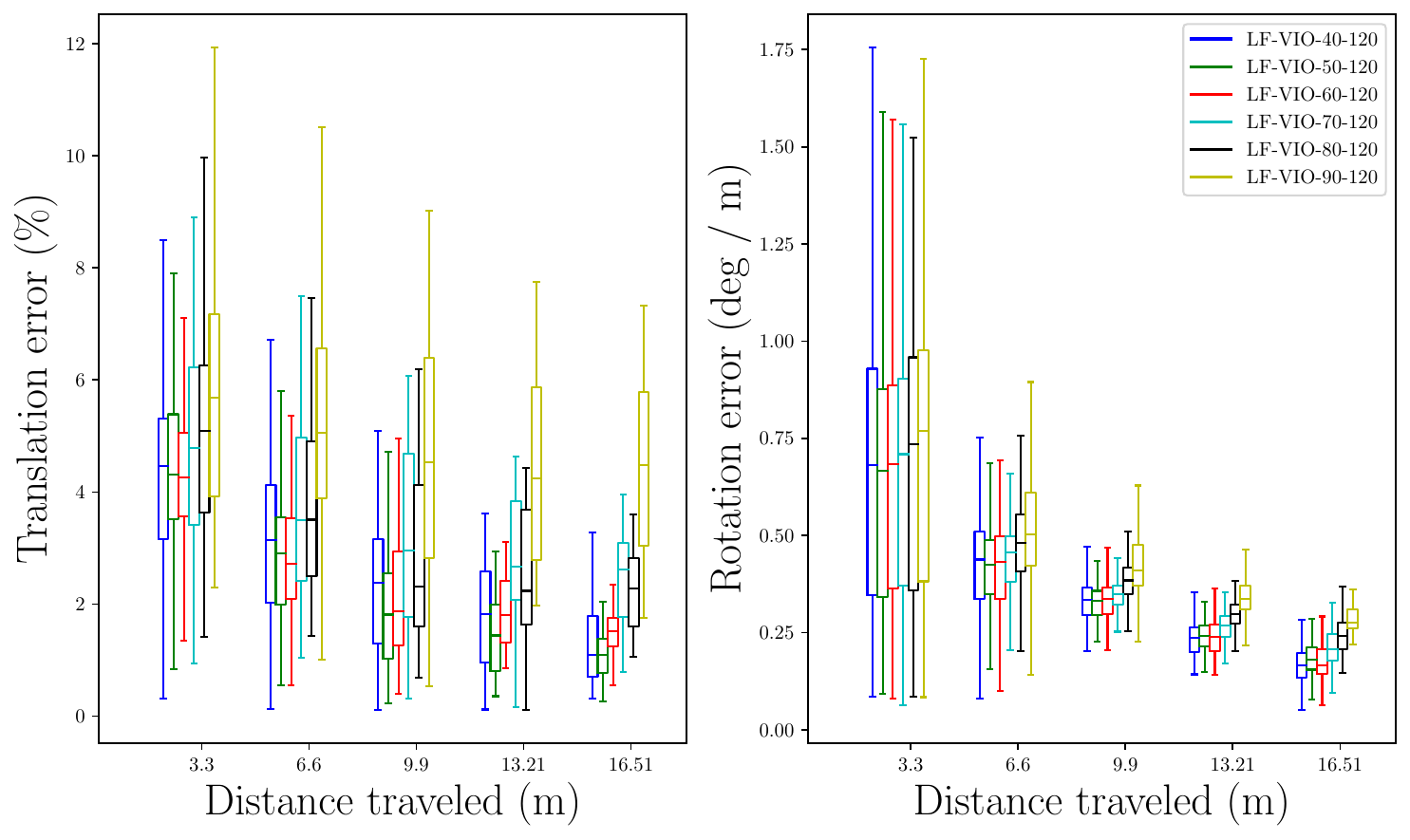}
		\label{fig:down_error}
	}
	\vskip-1ex
	\caption{Top trajectories and error analyses on the PALVIO ID06 set using different FoVs.}
	\label{fig:lfvio-field}
	\vskip-1ex
\end{figure*}

\begin{table*}[!t]
 \setlength{\tabcolsep}{2.2pt}
	\centering
	\begin{threeparttable}
	\caption{Comparison of VIO methods on the established PALVIO dataset.}
	\label{tab:b_lfvio_vio}
\renewcommand\arraystretch{1.4}{\setlength{\tabcolsep}{2.9mm}{\begin{tabular}{cccccccccccc}
\toprule   
\multicolumn{2}{c}{\multirow{2}{*}{VIO-Method}} & \multicolumn{10}{c}{Sequences}                                                                                                                                          \\
\multicolumn{2}{c}{}                            & ID01           & ID02           & ID03           & ID04           & ID05           & ID06           & ID07           & ID08           & ID09           & ID10           \\
\midrule
\midrule
\multirow{3}{*}{LF-VIO}      & RPEt (\%)         & \textbf{3.556} & \textbf{2.709} & \textbf{2.542} & \textbf{1.495} & \textbf{2.016} & \textbf{2.814} & \textbf{2.775} & 2.983          & 2.146          & \textbf{4.493} \\
                             & RPEr (degree/m)  & \textbf{0.328} & 0.599 & 0.292          & 0.227          & 0.328          & 0.397          & 0.315          & \textbf{0.202} & \textbf{0.322} & \textbf{0.567}          \\
                             & ATE (m)           & \textbf{0.341} & \textbf{0.153} & \textbf{0.269} & \textbf{0.166} & 0.200          & \textbf{0.093} & \textbf{0.237} & 0.236          & \textbf{0.222} & \textbf{0.292} \\
\midrule
\multirow{3}{*}{{LF-VIO-40-90}}      & {RPEt (\%)}        & {4.245} & {5.707} & {3.468} & {2.571} & {2.361} & {2.933} & {3.051} & {3.05} & {2.954} & {5.255} \\
                             & {RPEr (degree/m)}  & {0.405} & {0.892} & {0.370} & {0.316} & {0.378}          & {0.433}  & {0.368}  & {0.505} & {0.413} & {0.572}          \\
                             & {ATE (m)}           & {0.476} &  {0.583} & {0.401} & {0.312} & {0.327}          & {0.124} & {0.384} & {0.309} & {0.463} & {0.349} \\
\midrule
\multirow{3}{*}{SVO2.0~\cite{forster2014svo}}      & RPEt (\%)         & 6.531          & 6.995          & 2.710          & 1.928          & 2.354          & 3.409          & 3.718          & 2.811          & \textbf{2.012} & 14.147          \\
                             & RPEr (degree/m)  & 0.401          & \textbf{0.378}         & \textbf{0.235} & \textbf{0.165}          & 0.296          & \textbf{0.320} & \textbf{0.187} & 0.291          & 0.230          & 0.608 \\
                             & ATE (m)           & 0.761          & 0.380          & 0.366          & 0.174          & \textbf{0.148} & 0.124          & 0.428          & 0.236          & 0.292          & 1.122          \\
\midrule
\multirow{3}{*}{VINS-Mono~\cite{qin2018vins}}   & RPEt (\%)         & 5.446          & 3.542          & 2.767          & 2.189          & 2.553          & 2.993          & 2.941          & \textbf{2.405} & 2.933          & 4.494         \\
                             & RPEr (degree/m)  & 0.4577         & 0.605          & 0.285          & 0.249          & \textbf{0.278} & 0.445          & 0.339          & 0.452          & 0.457          & \textbf{0.567}         \\
                             & ATE (m)           & 0.870          & 0.214          & 0.310          & 0.217          & 0.263          & 0.104          & 0.299          & \textbf{0.194} & 0.378          & 0.557   \\ 
	
	\bottomrule
	\end{tabular}}}
	\end{threeparttable}
	\vskip-5ex
\end{table*}

Then, we experiment by decreasing the FoV from $50^{\circ}{\sim}120^{\circ}$ to $90^{\circ}{\sim}120^{\circ}$. 
In the worst case, when only the information from the negative plane is used ($90^{\circ}{\sim}120^{\circ}$), the VIO framework still works.
But when incorporating slightly more information from the positive plane, \textit{i.e.}, in the cases of $80^{\circ}{\sim}120^{\circ}$ and $70^{\circ}{\sim}120^{\circ}$, the performance recovers to the same level of that with pure positive plane.
This again confirms the importance of information lying on the negative plane and the effectiveness of our method in properly harvesting and exploiting the negative-plane features, which are unused in previous works.
Fig.~\ref{fig:lfvio-field} shows the top trajectories and error analyses based on the FoV variations, which illustrates the same trend.
{We note that the performance of the $50^{\circ}{\sim}120^{\circ}$ setting does not exceed that of $40^{\circ}{\sim}120^{\circ}$, which is due to the lowest density (the number of vectors per unit area) in $40^\circ{\sim}50^\circ$, so the use of information of this band may not improve the performance.}
 
\subsection{Comparison with the State-of-the-Art}
With the data captured by the motion capture system as the ground truth, we provide the accuracy of our LF-VIO on the PALVIO benchmark, compared with state-of-the-art visual-inertial-odometry methods SVO2.0~\cite{forster2014svo} and VINS-Mono~\cite{qin2018vins}, in Table~\ref{tab:b_lfvio_vio}, where the processing are conducted on a laptop with an R7-5800H processor.
{For fair comparisons, we use the camera model from Scaramuzza \textit{et al.} in both SVO2.0 and VINS-Mono. We also report the results of LF-VIO using only information from the positive plane ($40^{\circ}{\sim}90^{\circ}$).}
We use Relative Pose Error in translation (RPEt), Relative Pose Error in rotation (RPEr), and Absolute Trajectory Error (ATE) as the evaluation metrics for assessing the VIO's performances.

\begin{figure*}[t]
	\centering
	\subfigure[{Top Trajectory on ID01}]{
		\includegraphics[width=0.3\textwidth]{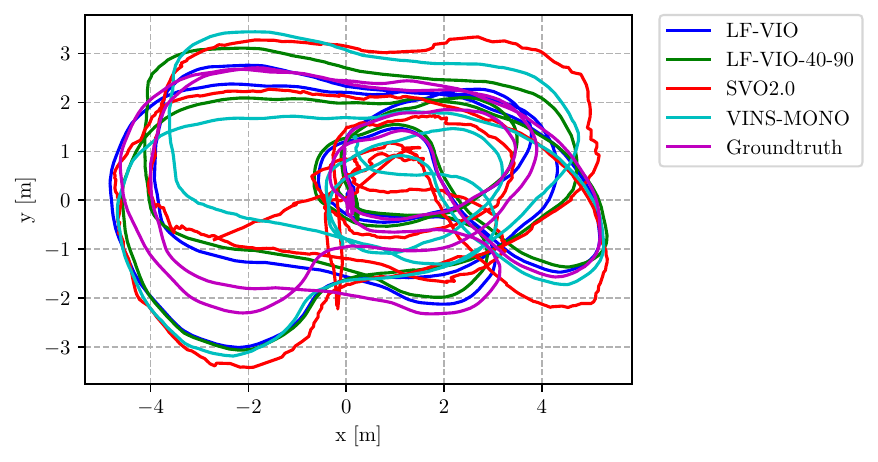}
		\label{fig:mh_01_trajectory_top}
	}
	\subfigure[{Top Trajectory on ID06}]{
		\includegraphics[width=0.3\textwidth]{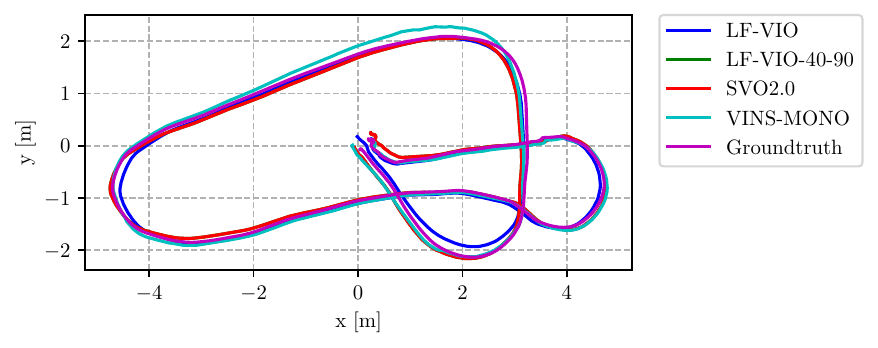}
		\label{fig:mh_06_trajectory_top}
	}
	\subfigure[{Top Trajectory on ID10}]{
		\includegraphics[width=0.3\textwidth]{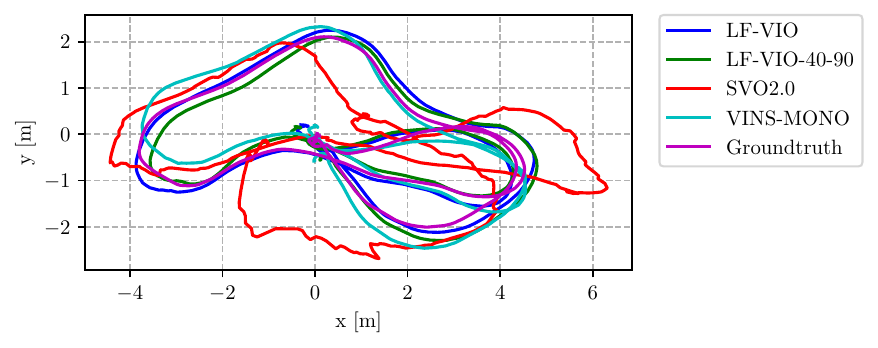}
		\label{fig:mh_11_trajectory_top}
	}
	\subfigure[{Translation and Rotation Error on ID01}]{
		\includegraphics[width=0.3\textwidth]{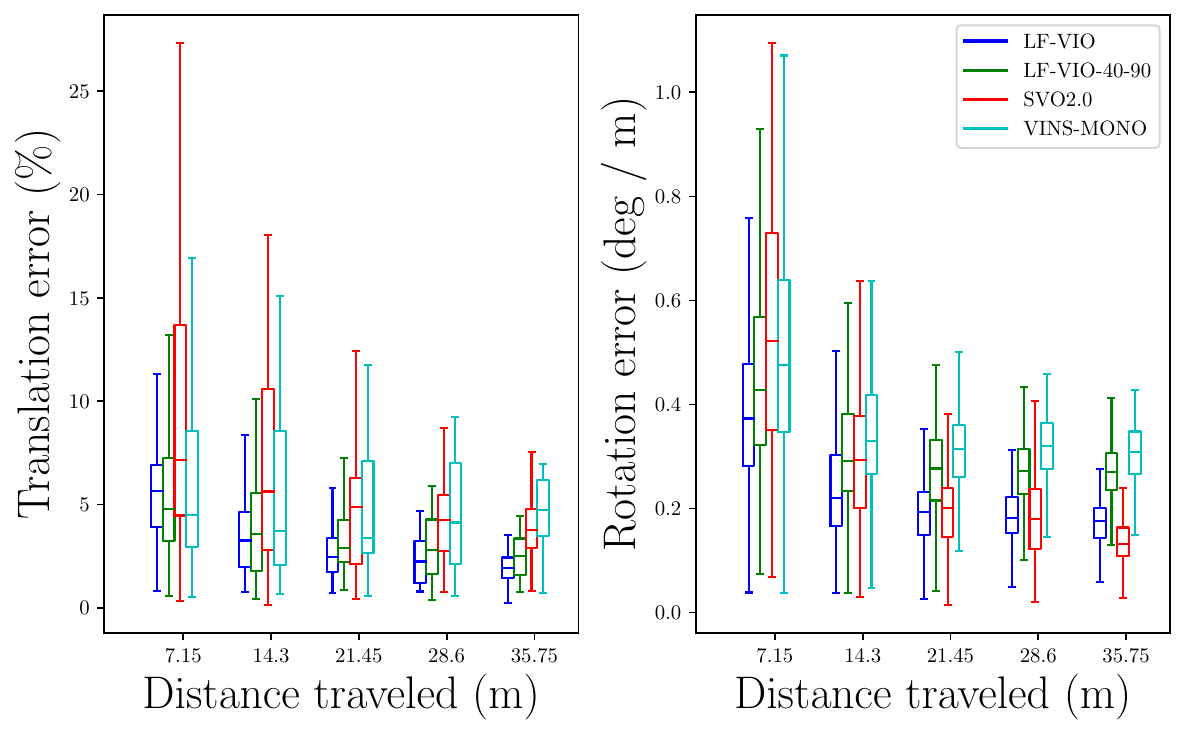}
		\label{fig:mh_01_trans_rot_error}
	}
	\subfigure[{Translation and Rotation Error on ID06}]{
		\includegraphics[width=0.3\textwidth]{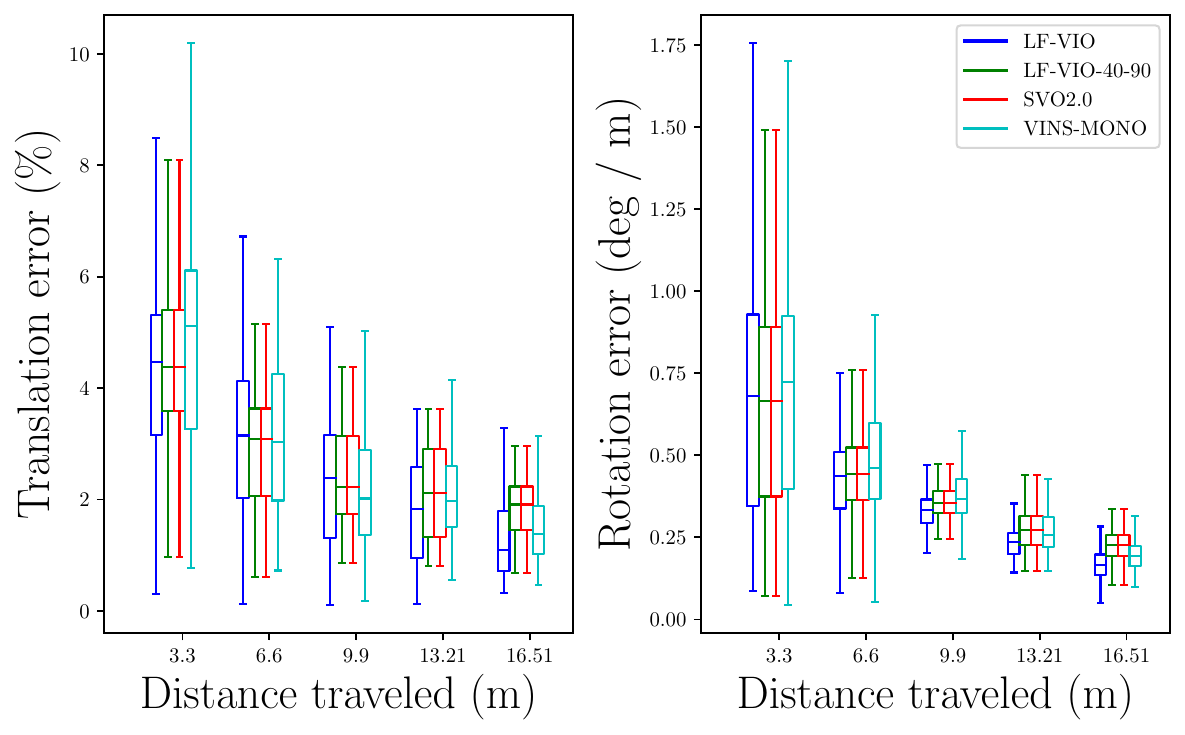}
		\label{fig:mh_06_trans_rot_error}
	}
	\subfigure[{Translation and Rotation Error on ID10}]{
		\includegraphics[width=0.3\textwidth]{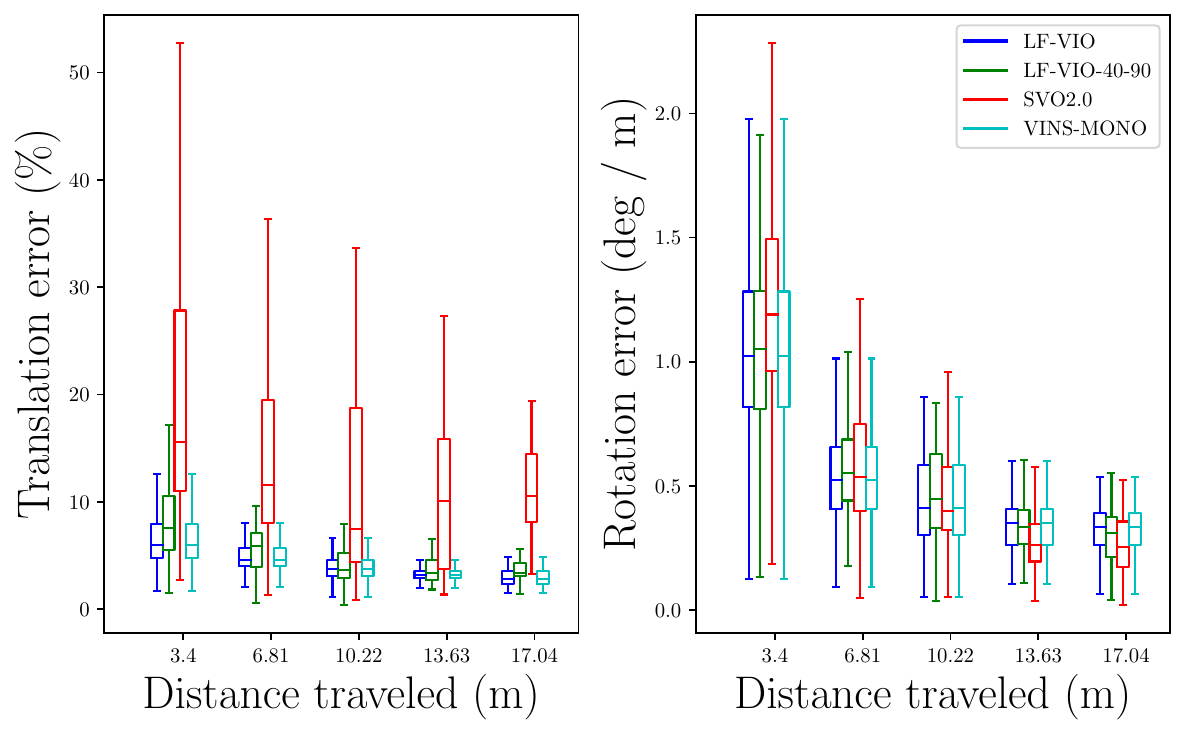}
		\label{fig:mh_11_trans_rot_error}
	}
	\vskip-1ex
	\caption{Examples of top trajectories and error analyses on the PALVIO benchmark of different VIO systems.}
	\label{fig:lfvio-vins-svo}
	\vskip-1ex
\end{figure*} 
It can seen that our method reaches the best precision in RPEt and ATE on most sequences, confirming the superiority of the proposal for VIO with large-FoV panoramic cameras.
{Using information from both positive- and negative plane improves the performance of LF-VIO.}
LF-VIO clearly stands out among the compared methods and yields accurate performances in almost all sequences, while the other two cannot always produce robust pose estimation. 
Further, we choose ID01, ID06, and ID10, as three typical and representative datasets for analysis in Fig.~\ref{fig:lfvio-vins-svo}.
In ID01 and ID10, the maximum Z-axis angular velocity reaches to $2.1$ and $2.3$ rad/s, and the images begin to become blurry.
Thus, the accuracy of all the algorithms are less satisfactory.
However, our method shows the relatively high accuracy in RPEt, RPEr, and ATE. 
In ID06, the maximum Z-axis angular velocity only reaches $1.5$ rad and the ATE of all the compared algorithms are rather low.
Moreover, in ID10, SVO2.0 has a poor performance under high angular velocity, due to unstable feature points and scale estimation in the back-end, but our system still maintains robust.
Generally, our method has a higher accuracy and robustness in situations which have relatively fast rotation and it also has a good performance in low-rotation scenarios.
In summary, our approach is verified to be suitable for wide-FoV panoramic cameras.

\subsection{Generalization to LiDAR-Visual-Inertial Odometry} 
The proposed method supporting negative half plane can also be transplanted to and integrated with LiDAR-Visual-Inertial Systems, such as LVI-SAM~\cite{shan2021lvi}.
We compare the accuracy of this system with and without our proposed method in Table~\ref{tab:b_lf_lvi_sam} on the two sequences, \textit{i.e.}, $\emph{Handheld}$- and $\emph{Jackal}$ datasets, provided by~\cite{shan2021lvi}, where the results are obtained via a laptop with an R7-5800H processor.
Their recording camera has a FoV of $360^\circ{\times}(0^\circ{\sim}93.5^\circ)$.

\begin{figure}[t!]
	\centering
	\includegraphics[width=1.0\linewidth]{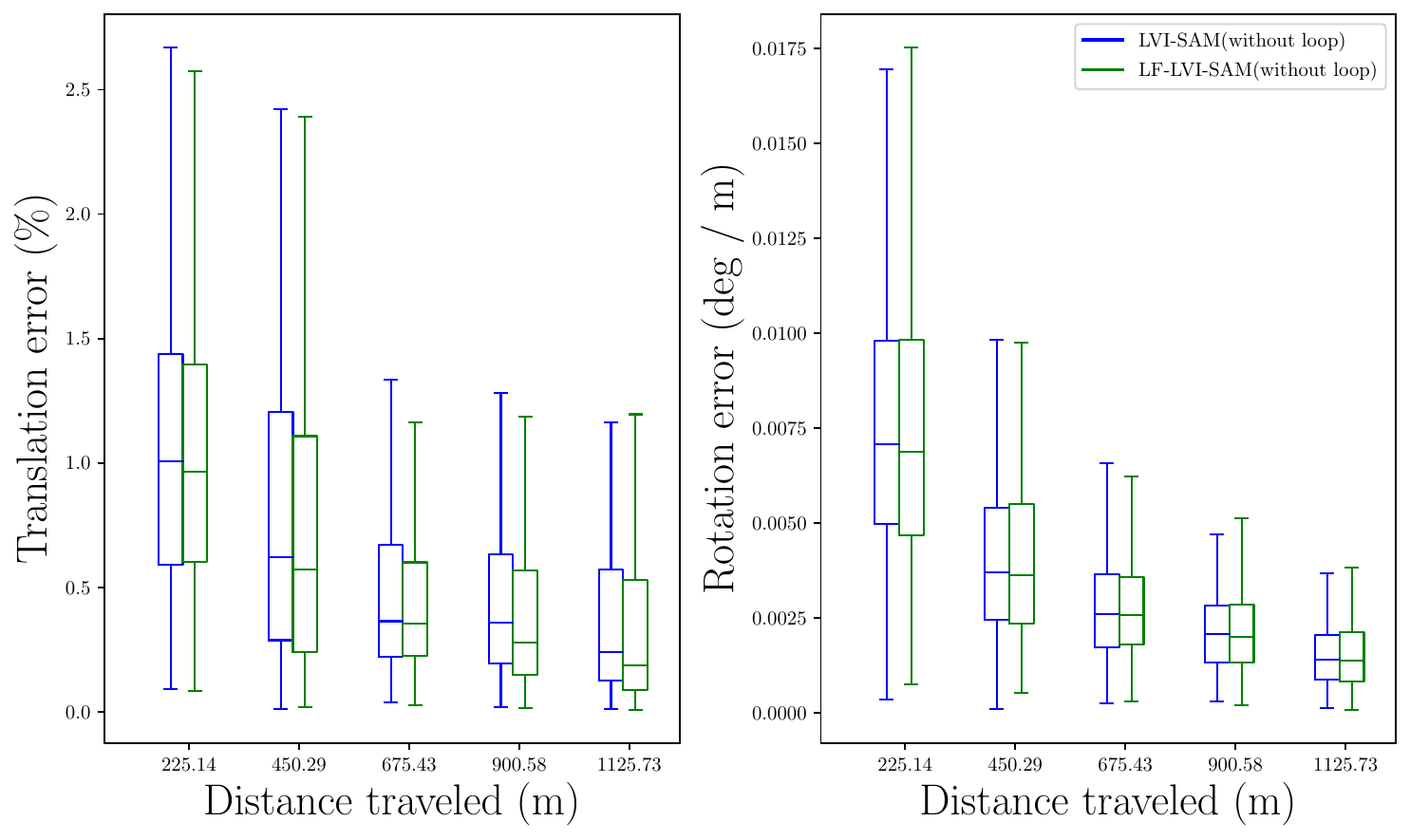}
	\vskip-2ex
	\caption{Translation and Rotation Error on the Handheld set.}
    \label{fig:lvisam_h}
    \vskip-2ex
\end{figure}

\begin{figure}[t!]
	\centering
	\includegraphics[width=1.0\linewidth]{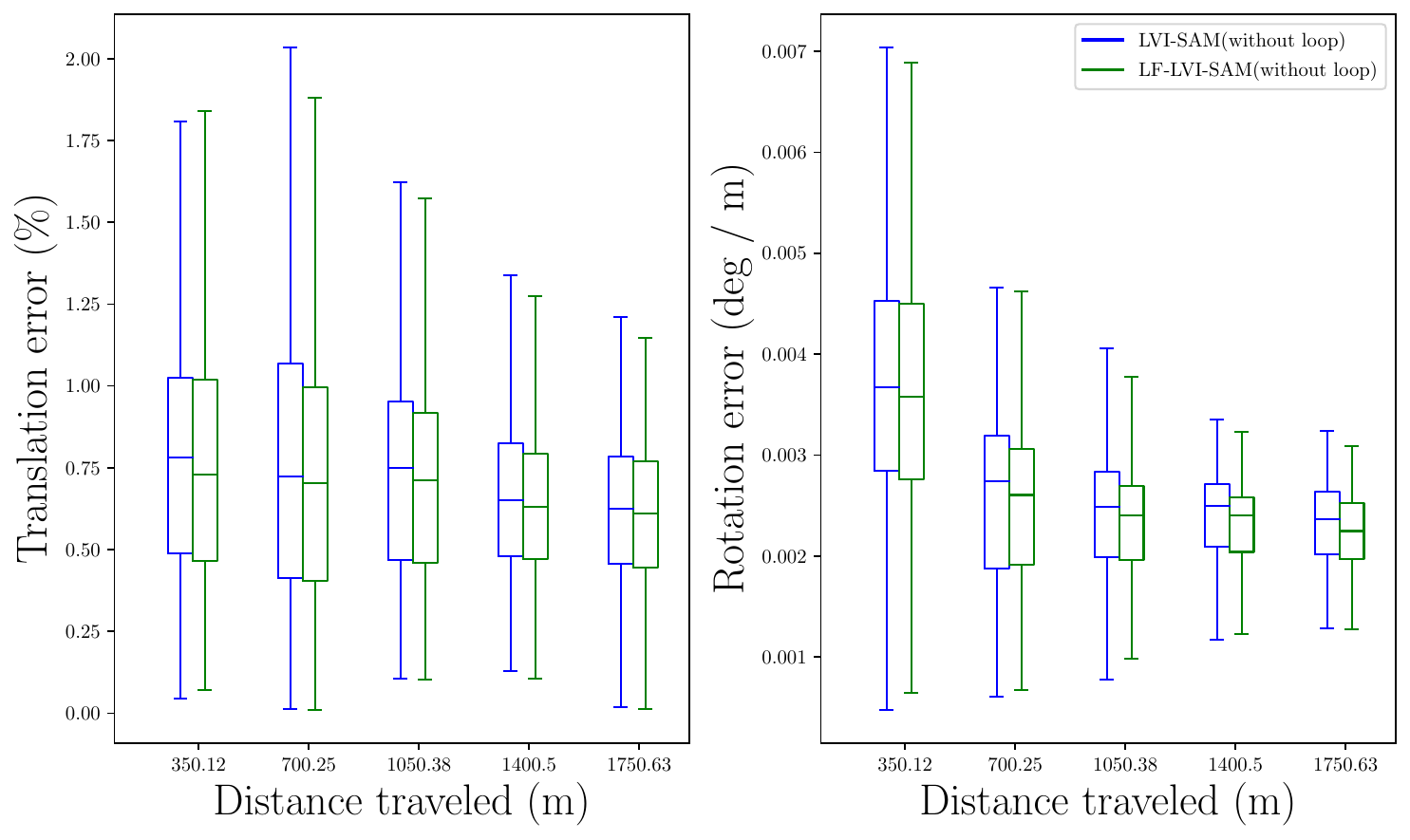}
	\vskip-2ex
	\caption{Translation and Rotation Error on the Jackal set.}
    \label{fig:lvisam_j}
    \vskip-3ex
\end{figure}

\begin{table*}[!t]
 \setlength{\tabcolsep}{2.2pt}
	\centering
	\begin{threeparttable}
	\caption{Comparison of VIO methods on the LVI-SAM dataset~\cite{shan2021lvi} (without  loop).}
	\label{tab:b_lf_lvi_sam}
	\renewcommand\arraystretch{1.4}{\setlength{\tabcolsep}{0.95mm}{\begin{tabular}{cccccccccccccc}
    \toprule    
    \multirow{2}{*}{Dataset}  & \multirow{2}{*}{LVIO-Method}      & \multicolumn{4}{c}{RPEt(\%)} & \multicolumn{4}{c}{RPEr (degree/m)} & \multicolumn{4}{c}{ATE(m)} \\
                          &                                  & Mean   & Min  & Max  & RMSE  & Mean    & Min    & Max    & RMSE    & Mean  & Min  & Max  & RMSE \\ 
    \midrule
    \midrule
    \multirow{2}{*}{$\emph{Handheld}$} & LF-LVI-SAM (Ours) 	& \textbf{0.63857} &  \textbf{0.00660}&4.38248&\textbf{0.84366}&
    \textbf{0.00393}&\textbf{0.00007}&0.02180& \textbf{0.00510}&
    \textbf{8.89108}&\textbf{0.08698}&\textbf{25.30987}&\textbf{11.57598}
	\\
    & LVI-SAM~\cite{shan2021lvi}    	&0.67193&0.01113&\textbf{4.37260}&0.87484&
	0.00397&0.00010&\textbf{0.01947}&0.00511&
	10.78413&0.09040&26.72794&13.52003
	\\
    \midrule 
    \multirow{2}{*}{$\emph{Jackal}$} & LF-LVI-SAM (Ours) & \textbf{0.70924} & \textbf{0.00948}&2.83097& \textbf{0.79693}&
	\textbf{0.00265}&\textbf{0.00047}&\textbf{0.00761}&\textbf{0.00283}&
	\textbf{5.80506}&0.80519&\textbf{14.21643}&\textbf{6.28478}
	\\
    & LVI-SAM~\cite{shan2021lvi} 	
    &0.73586&0.01300&\textbf{2.80597}&0.82791&
	0.00274&\textbf{0.00047}&0.00781&0.00291&
	6.02435&\textbf{0.65769}&14.53112&6.53074
	\\  
	
	\bottomrule
	\end{tabular}}}
	\end{threeparttable}
	\vskip-3ex
\end{table*}
In this experiment, we only change the visual part of LVI-SAM.
We use a unit vector to represent feature points and adjust visual algorithms about it {and the MEI camera model~\cite{mei2007single} following the original setup~\cite{shan2021lvi}}.
LVI-SAM~\cite{shan2021lvi} uses a mask that has many large field area, in order to avoid the problems coming with the feature points on the negative plane or those close to $90^\circ$.
In our system, this challenge is solved, so we reduce the area of the mask to explore the best performance of our system.
It can be seen that LF-LVI-SAM is consistently better than LVI-SAM in Mean and RMSE of RPE (\%), RPEr (degree/m) and ATE (m). 
The RPE on the $\emph{Handheld}$- and $\emph{Jackal}$ datasets are shown in Fig.~\ref{fig:lvisam_h} and Fig.~\ref{fig:lvisam_j}. 
In summary, our method can not only improve the precision of VIO systems, but also be effective for LiDAR-VIO systems whose FoV reaches the negative half plane.

\subsection{Speed Analysis}
We report the efficiency of our approach in this subsection, which mainly composed of the front-end and the back-end.
If the mapping relationship of the images is stored in memory, the image mapping process takes about $1{\sim}2ms$.
In the front-end of LF-VIO, this operation is not necessary, but the feature tracking and optical flow computation cost $15ms$ and $3ms$ respectively, and the total front-end costs about $40ms$.
In the back-end, the solver costs around $30ms$ mainly and the total costs about $60ms$. Besides, the frond-end and back-end operate independently.
In general, our LF-VIO system can reach at least $10Hz$ on the onboard computer with a quad-core Intel i7-8550U processor, which is reasonable for real-time mobile applications.

\section{Conclusions}
In this paper, we have proposed \emph{LF-VIO}, a framework for large-FoV cameras with a negative plane, for performing real-time pose estimation.
The proposed framework is composed of two parts: a robust initialization process and a tightly coupled optimization process.
We create and release the \textit{PALVIO} dataset with $10$ sequences collected via a Panoramic Annular Lens (PAL) camera and an IMU sensor, for evaluating the performance of our proposed framework and fostering future research on panoramic visual inertial odometry.
Our method improves the system robustness and accuracy, outperforming state-of-the-art VIO approaches.
Our method generalizes well to other SLAM systems which involve large-FoV sensing and the feature point method, {as} proved when it is combined with LVI-SAM.
Our implementation has been open sourced.
In the future, we are interested in adding loop closure and combing the stereo panoramic information and LiDAR sensors in our system.

\bibliographystyle{IEEEtran}
\bibliography{bib}

\end{document}